%% file: main.tex
\documentclass{article}
\PassOptionsToPackage{numbers, square}{natbib}

\usepackage[final]{neurips_data}

\usepackage[utf8]{inputenc} 
\usepackage[T1]{fontenc}    
\usepackage{hyperref}       
\usepackage{url}            
\usepackage{booktabs}       
\usepackage{amsfonts}       
\usepackage{nicefrac}       
\usepackage{microtype}      
\usepackage[table,xcdraw]{xcolor}         

\usepackage{multirow}
\usepackage{longtable}
\usepackage{booktabs}
\usepackage{sidecap}
\usepackage{amsthm}
\usepackage{amsmath}
\usepackage{amssymb}
\usepackage{amsfonts}
\usepackage{verbatim} 
\usepackage{url}
\usepackage{pifont}
\usepackage{eqparbox}
\usepackage{arydshln} 
\usepackage{bigstrut}
\usepackage{comment}

\usepackage[linewidth=1pt]{mdframed}
\usepackage{framed}
\usepackage{xcolor,colortbl}
\usepackage{pgffor}
\usepackage{comment}
\usepackage{graphicx}
\usepackage{booktabs}
\usepackage{subcaption}
\usepackage{enumitem}
\usepackage{makecell}
\usepackage{listings}
\usepackage{cleveref}
\usepackage{soul}

\usepackage{boxedminipage}
\usepackage{xspace}
\usepackage{array}
\usepackage{framed}
\usepackage{caption}
\usepackage{tikz}
\usetikzlibrary{tikzmark}
\usepackage{enumitem}
\usepackage{bigstrut}
\usepackage{placeins}
\usepackage[notextcomp]{stix}
\usepackage{wrapfig}
\usepackage{stackengine}

\newcommand{\eg}{e.g.,\xspace}

\def\tabref#1{Table~\ref{tab:#1}}

\def\tablabel#1{\label{tab:#1}\label{p:#1}}
\def\eqref#1{Eq.~\ref{eqn:#1}}

\def\Secref#1{Section~\ref{sec:#1}}
\def\Appref#1{Appendix~\ref{sec:#1}}
\def\seclabel#1{\label{sec:#1}\label{p:#1}}
\def\applabel#1{\label{sec:#1}\label{p:#1}}
\def\secref#1{\S\ref{sec:#1}}

\newcommand{\huggingface}{\raisebox{-1.5pt}{\includegraphics[height=1.05em]{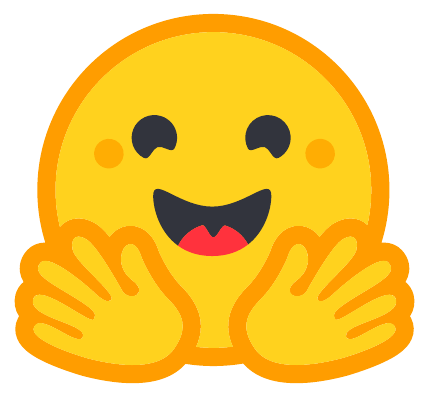}}\xspace}
\newcommand{\github}{\raisebox{-1.5pt}{\includegraphics[height=1.05em]{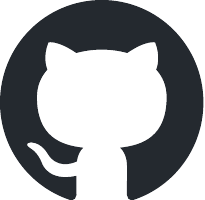}}\xspace}

\def\genericname{\mbox{{GlotCC}}\xspace}
\def\glotlid{\mbox{{GlotLID}}\xspace}
\def\cc{\mbox{{CommonCrawl}}\xspace}

\def\datasize{2TB\xspace}
\def\numlanguages{1000\xspace}
\def\numlanguagesexact{1275\xspace}

\def\fasttext{\mbox{FastText}\xspace}

\title{\genericname: An Open Broad-Coverage \cc Corpus and Pipeline for Minority Languages}

\author{
    Amir Hossein Kargaran$^{\clubsuit}$ \; \; 
    François Yvon$^{\spadesuit}$ \; \; 
    Hinrich Schütze$^{\clubsuit}$ \\
    \\
    $^\clubsuit$LMU Munich \& Munich Center for Machine Learning, Munich, Germany  \\
    $^\spadesuit$Sorbonne Université \& CNRS, ISIR, Paris, France \protect \\
    \texttt{amir@cis.lmu.de}
}

\begin{document}

\maketitle

\begin{abstract}

The need for large text corpora has increased with the advent of pretrained language models and, in particular, the discovery of scaling laws for these models. Most available corpora have sufficient data only for languages with large dominant communities. However, there is no corpus available that (i) covers a wide range of minority languages; (ii) is generated by an open-source reproducible pipeline; and (iii) is rigorously cleaned from noise, making it trustworthy to use. We present \genericname, a clean, document-level, \datasize general domain corpus derived from \cc, covering more than \numlanguages languages. We make \genericname and the system used to generate it— including the pipeline, language identification model, and filters—available to the research community.

\begin{center}
\renewcommand{\arraystretch}{1.2}
\footnotesize
\begin{tabular}{rccl}
     \huggingface & \textbf{Corpus}&{\small\texttt{v.\:1.0}} & \href{https://huggingface.co/datasets/cis-lmu/GlotCC-v1}{\path{hf.co/datasets/cis-lmu/GlotCC-v1}}\\
     \github & \textbf{Pipeline}&{\small\texttt{v.\:3.0}} & \href{https://github.com/cisnlp/GlotCC}{\path{github.com/cisnlp/GlotCC}} \\
\end{tabular}
\end{center}

\end{abstract}

\section{Introduction \seclabel{introduction}}

Progress in building language technologies has largely been
limited to about 200 languages
(referred to as ``large languages'' in this paper)
for which text
is reasonably accessible~\citep{joshi-etal-2020-state}. In
this regard there is a
disparity between large languages and minority languages, a
disparity that
becomes much more visible as state-of-the-art language
technologies
-- developing from word embedddings and encoder-only models
to large autoregressive models --
become more data-hungry.  FastText~\citep{bojanowski-etal-2017-enriching, grave2018learning} word
vectors supports $\approx$160 languages,
XLM-R~\citep{conneau-etal-2020-unsupervised} 
$\approx$100 languages, 
Llama-3~\citep{meta2024introducing}  $\approx$30
languages. So there is a trend for the
number of supported
languages to decrease over time in state-of-the-art
models. 

While efforts have been made to compile multilingual corpora 
from books~\citep{leong-etal-2022-bloom,
  gao2020pile},  the most common approach to collecting
large amounts of raw textual data is to rely on
crawled web text~\citep{nllbteam2022language,
  abadji2022towards, bapna2022building, kudugunta2024madlad,
  soldaini2024dolma}. However, there are still many issues
to resolve to guarantee the quality of the extracted web
text, particularly for minority
languages~\citep{kreutzer2022quality}.
Most pipelines for
web-text extraction rely on language identification (LID) to
classify text,
e.g., on 
FastText \citep{bojanowski-etal-2017-enriching, joulin2016fasttextzip} (as does 
OSCAR \citep{abadji2022towards})
or
on CLD3
\citep{botha-etal-2017-natural,
  salcianu2018compact,xue-etal-2021-mt5}.
Many
errors in existing web corpora are
artifacts of their LIDs \citep{caswell-etal-2020-language}.
For example, FastText's accuracy suffers from hash collisions:
it maps out-of-domain (OOD) n-grams
to n-grams seen in training, resulting in many
OOD errors.
 Additionally, commonly used LIDs
can identify the (roughly) 200 largest languages but only a few
minority languages, leading to  ``out-of-model
cousin'' errors \citep{caswell-etal-2020-language,
  kreutzer2022quality}.

In this paper, we adopt the Ungoliant pipeline
\citep{abadji2021ungoliant} for extracting text
from \cc.
To address the limitations of current LID models (hash
collisions and limited language coverage),
we develop a new LID model, \glotlid v3.0, an
extension of 
\glotlid v1.0
\citep{kargaran-etal-2023-glotlid}. This model covers over
2,000 linguistic labels. It also mitigates common
sources of noise encountered on the web with a better rejection model.
We also extend
Ungoliant with several
filtering techniques
that remove
general web
noise, list-like content, documents with repeated words
\citep{kudugunta2024madlad, rae2021scaling}
and ``inconsistent'' documents, i.e., documents for which 
LID detects multiple languages, often an indicator of noise.
We perform an audit of random 653 language subcorpora of \genericname, finding
that the data is  in-language, with a
macro-average score of 0.93 and median score of 1.0. 
We publish \genericname as a 
document-level corpus.
This makes it usable
for pretraining generative
language models as well as for other language
technologies that require information beyond the sentence level.

\section{\glotlid \seclabel{glotlid}}

In our previous work, we introduced \glotlid
v1.0~\citep{kargaran-etal-2023-glotlid}, an LID model based
on the \fasttext architecture
\cite{bojanowski-etal-2017-enriching} with good performance
for 1665 languages.
Upon further analysis, we improved
\glotlid
to v2.0 and then v3.0. In v2.0, we
extended our ISO 639-3 labels to also include 
the ISO-15924 script;
an example is rus-Cyrl
(Russian language in Cyrillic script).
This new labeling
reduces errors  and enhances the quality of
language resources. For instance, it allows us to restrict the
set of language labels that can be assigned to each script,
thereby avoiding obvious mismatches between language and
script \citep{kargaran-etal-2024-glotscript-resource}. This
can be implemented by employing writing system detection
tools, such as GlotScript-T
\citep{kargaran-etal-2024-glotscript-resource}, during
inference. We also use GlotScript-T and GlotScript-R to
filter out noisy training data from the \glotlid training corpus. GlotScript-T
determines the script of an input text. GlotScript-R
provides the admissible script(s) for each language.

In \glotlid v3.0, we increase the number of
LID labels to more than 2000 by adding, removing and
relabeling resources based on feedback from the
community and on our own findings from the continuous
analysis we are 
conducting.

\subsection{\glotlid v3.0 vs \glotlid v1.0/v2.0}

\subsubsection{Increased coverage of minority languages}\seclabel{new_langs}

We include in the \glotlid training set new resources for
African languages~\citep{lastrucci-etal-2023-preparing,
  siminyu2021ai4d, Adelani2023MasakhaNEWS,
  ogundepo2023afriqa}, Uralic
languages~\citep{jauhiainen2019wanca,
  yankovskaya-etal-2023-machine}, Indonesian
languages~\citep{winata2022nusax}, Indic
languages~\citep{madhani2023bhashaabhijnaanam,
  gala2023indictrans} as well as additional indigenous
languages~\citep{bustamante-etal-2020-data}.  We also
consider data for a variety of languages from
books~\citep{leong-etal-2022-bloom}, children
storybooks~\citep{GlotStoryBook},
crowdsourcing~\citep{ardila-etal-2020-common}, low-resource
crawls~\citep{GlotSparse},
and world language
repositories~\citep{nivre-etal-2020-universal}.

\subsubsection{Better rejection model with the ``UND'' label}\seclabel{und}

LID is typically understood as a closed-set classification
problem; most LIDs, including \glotlid{}, adopt this setup.
As LID is in fact an open-set problem
\cite{jauhiainen-etal-2018-automatic, malmasi2017open}, when
processing web data, there is always the risk of encountering
``unknown'' languages (those not occurring in train). To limit  errors
that are caused by these unknowns, it is customary to reject
samples for which the most likely language has a low
probability, assuming the classification model is
well-calibrated \cite{vaze-etal-2022-openset,
  kargaran-etal-2023-glotlid}.

This problem is compounded when using popular LID
architectures such as \fasttext and CLD3, which detect
languages based on character n-gram features. To limit
memory footprint and  speed up inference, \fasttext{}
hashes character n-grams into a predefined range
of integers, using this map to retrieve 
feature embeddings.
This procedure applies to  utf-8
character $n$-grams, where the default
for \glotlid is $2\leq n \leq 5$.

This approach  makes
no distinction  between
n-grams that have been seen in training and those that have not, as any
n-gram is assigned to an existing hash
value. Furthermore, as the number of languages increases, so
does the number of scripts, and accordingly, the number of
possible n-grams. With 160+ scripts, this number is in fact
much greater than \fasttext's default hash size (2,000,000), increasing
the chance of collisions. This first implies that the
probability of languages that are well represented in the
training data (i.e., high-resource languages) or that have a
large number of n-grams due to their character set
(e.g., Chinese) is overestimated. This also means
that closed-set LIDs will provide a non-zero probability
score even for writing systems that were never seen during
training, predicting (sometimes with high
probability) languages whose n-grams
collide with unseen n-grams.

Although \glotlid supports all major scripts, its
training data does not contain minor 
scripts such as Mong (Mongolian), Sylo (Syloti Nagri), Newa
(Pracalit), Talu (New Tai Lue) and Gran (Grantha). In order to
correctly reject languages written in these scripts, and  
given that we have no access to reliable
training data for them,
we introduce a set of 157 new ``und'' 
(\emph{undetermined}) labels. The associated training data is
obtained by randomly generating character strings from the
corresponding character set. For example, for the Talu
script, we use the label
``und\_Talu''. ``und\_Talu'' training data consists of randomly
selected Talu characters forming 100,000~sentences.

\subsubsection{Removing noise with ``zxx'' labels}
\seclabel{zxx}
In addition to handling unseen scripts, web crawlers and
processing systems also need to be robust to multiple types
of noise \citep{caswell-etal-2020-language}.
To make \glotlid
more robust, we create training data for
additional types of noise, including
mis-rendered PDFs and Mojibake (text
decoded using an unintended character encoding).
We either create artificial training data or we
use an initial seed and query the Google
search engine to obtain training data.

We include six major sources of web noise that we
encountered in our work on \glotlid.

1) \textbf{Mis-rendered PDF:} This is gibberish Latin text
that appears when an Arabic PDF is mis-rendered by OCR.
To generate representative sentences,
we query 1- and 2-grams
of the letters 'i', 'j' and 'l', with
spaces between them, as these letters are
frequent in this
context.\footnote{{Mis-rendered PDF query example:} \href{https://www.google.com/search?q=i+j+l+ii+jj+ll+ij+ji+lj+jl+li+il}{\path{google.com/search?q=i+j+l+ii+jj+ll+ij+ji+lj+jl+li+il}}}

2) \textbf{A N T S P E A K:} This is a type of noise where
the characters of the text are
space-separated~\citep{caswell-etal-2020-language}.
It
is easy to generate synthetic training data from any text or alphabet.

3) \textbf{Binary files:} This type of noise occurs when
binary files, especially images,
become part of the text.
It
is easy to generate training data by
opening any binary file in a
text editor.

4) \textbf{Mojibake Latin:} This is gibberish text that
results from text being decoded using an unintended Latin
character encoding. In this type of noise, vowel characters
of latin with different accents in a repeated form can serve
as a seed for Google queries, such as
``áàãà''.\footnote{{Mojibake Latin query example:} \href{https://www.google.com/search?q=áàãà}{\path{google.com/search?q=áàãà}}}

5) \textbf{Mojibake Arabic:} This is gibberish text that
results from text being decoded using an unintended Arabic
character encoding. In this type of noise,
ˆ, ®,
±,  § and
the Arabic
characters Tah and Zah appear 
frequently.
Our queries are based on these characters.\footnote{Mojibake Arabic webpage example: \href{https://web.archive.org/web/20200120063531/https://www.al-jazirah.com/2010/20100802/index.htm}{\path{al-jazirah.com/2010/20100802/index.htm}}}

6) \textbf{Replacement character:} The replacement character (\texttt{U+FFFD}) appears
repeatedly in this type of text. We simulate it by
randomly replacing characters with the replacement
character.

We create three \glotlid labels for these six noise types:
zxx\_Latn for noise types [1-4], zxx\_Arab for noise
type 5 and zxx\_Zzzz for noise type~6.

\subsubsection{Curation of labels}\seclabel{changed_labels}
To curate our set of LID labels, we rely on confusion
matrices, focusing on languages with low performance, i.e.,
those that are frequently confused with others.  We employ
both genealogical analysis and basic sanity checking using
web resources about the languages.
See Section~7
of ~\citep{kargaran-etal-2023-glotlid} for an example of our
methodology.
Most of the problems identified this way
were due to the training data, e.g.,  the training corpus
for a given language is actually a mix of several languages.
We remove labels and their training
data if we deem them too noisy based on our analysis.

\glotlid v1.0 and v2.0 both support
macro languages and individual languages. In cases where the correct
class is an individual language,
the associated probability is often spread over
this language and its macro language.
The reason is that many individual language n-grams and words 
are also frequent in the
macro language. 
In \glotlid v1.0/v2.0, 
we provide the option of
taking the softmax on a subset of LID labels (e.g.,
on just the individual language and the macro language).
However, according to community feedback, it is
preferable for  labels to be mutually exclusive,
making it possible to run LID just
once over all supported labels. We rectify this problem by
(i)
re-labeling macro languages as one of the individual
languages; (ii) merging the individual
languages into the macro language (in case of small differences between
them, e.g., we merge
individual languages prs\_Arab and  pes\_Arab
into macro language  fas\_Arab); and (iii)
deleting the macro language in case we
already have good support for its individual languages such as ``zho\_Hani'', ``aze\_Latn'', ``est\_Latn''.
However, there are some conventional exceptions: we keep both ``srd\_Latn'' and ``sdc\_Latn''. ``srd\_Latn'' only represents ``sro\_Latn'' and ``src\_Latn'', not ``sdc\_Latn''. This decision is based on the Glottolog tree~\citep{hammarstrom2024glottolog} and the linguistic map of Sardinia.  
We changed ``tgl\_Latn'' to ``fil\_Latn'' following \citet{kudugunta2024madlad}.

\subsection{Evaluation setup \seclabel{evaluation}} 

We train \glotlid v3.0 using the same parameters and
sampling strategy as \glotlid
v1.0~\citep{kargaran-etal-2023-glotlid}
(also described in \tabref{hyperparameters}) 
for 1 epoch.  In our previous work~\citep{kargaran-etal-2023-glotlid}, we found that the variance of the \fasttext architecture with different initial seeds is negligible.

\input{tables/glotlid_hyperparams}

\glotlid is trained on the \glotlid v3.0 corpus, which draws
from many different sources; the three main sources are
Wikipedia, news websites, and religious
texts.\footnote{\href{https://github.com/cisnlp/GlotLID/blob/main/sources.md}{\path{github.com/cisnlp/GlotLID/blob/main/sources.md}}}
The new languages (\secref{new_langs}), und labels
(\secref{und}), and zxx labels (\secref{zxx}) are also
included. As some of our evaluation benchmarks
(\secref{lid_evaluation_data}) might have leaked into other
sources included in the GlotLID corpus, such as UDHR in
articles (e.g., Wikipedia), translation community resources
(e.g., Tatoeba), and news (e.g., BBC), we remove this
contamination from the \glotlid corpus. We count a benchmark test sentence as occurring in the \glotlid corpus if all of its word four-grams occur in one sentence of the \glotlid corpus. We remove all of the sentences from the \glotlid corpus that meet this condition. After deduplication, we split the corpus into training, validation, and test sets in the ratio 0.8/0.1/0.1.

Following prior work \citep{nllbteam2022language,
  kargaran-etal-2023-glotlid, burchell-etal-2023-open},
we
report F1 and false positive rate (FPR)
and
assume that the set of languages covered by
the evaluation benchmark is known. Accordingly, we restrict a
model's predictions to those languages that occur in the
intersection of the benchmark and model training data.

\subsubsection{Evaluation data}\seclabel{lid_evaluation_data}

We evaluate \glotlid v3.0 on  three existing benchmarks with a
high  number of languages: UDHR,
FLORES-200 and GlotTest (our in-domain test set).

1) \textbf{GlotTest:} 
Let $n_l$ be the number of sentences from
language $l$ in the \glotlid corpus test set. Then we sample min(1000,
$n_l$) sentences from it. We refer to the resulting dataset
as GlotTest. GlotTest supports 2102 LID labels, the same number
of labels supported by \glotlid v3.0. This includes the
``und'' labels (\secref{und}) and ``zxx'' labels (\secref{zxx}).

2) \textbf{UDHR:} UDHR consists of about 500 translations of the
“Universal Declaration of Human Rights”.
In this work, we use
the UDHR
test set released in our previous
work~\citep{kargaran-etal-2023-glotlid}.\footnote{\href{https://huggingface.co/datasets/cis-lmu/udhr-lid}{\path{hf.co/datasets/cis-lmu/udhr-lid}}}
It supports the 415 translations from
\url{udhrinunicode.org} that are available
with a valid ISO 639-3 code. 371 of these 415
are covered by
\glotlid v3.0 training data.

3) \textbf{FLORES-200:}
FLORES-200~\citep{nllbteam2022language} comprises 842
articles sourced from English-language Wikimedia
projects. Each sentence of these articles was translated
into 204 language\_script labels. The dataset is split into
997 sentences for development, 1012 for dev-test and 992
for test. FLORES-200 test  is not
public. As is common practice,
we use FLORES-200 dev-test  as our FLORES-200 test
set.

\subsubsection{Evaluation results}

\tabref{lid_performance} reports results on GlotTest, UDHR
and FLORES-200. On average, \glotlid v3.0 achieves an F1
score of 0.991 and a false positive rate of 0.000003 on
GlotTest.  The three lowest F1 scores on GlotTest are 0.72,
0.75, and 0.76 for bos\_Latn, hrv\_Latn, and cnr\_Latn,
three mutually intelligible varieties of BCMS
(Bosnian-Croatian-Montenegrin-Serbian). We made sure that
every LID label has performance of at least 0.7 on
the \glotlid validation set. Otherwise, we remove or merge
the label (see \secref{changed_labels}).

\input{tables/glotlid_abs}

Compared to \glotlid
v1.0
\citep{kargaran-etal-2023-glotlid},
\glotlid v3.0 shows improvements in F1 of 0.05 for GlotTest, 0.09
for UDHR, and 0.05 for FLORES-200. Although the UDHR results are the lowest among these
three benchmarks (F1 of 0.882 vs 0.991 and 0.967), \glotlid
v3.0 outperforms the state-of-the-art also for this
dataset~\citep{kargaran-etal-2023-glotlid}. The most likely
reason for the lower performance on UDHR is a
domain shift:
the data in the \glotlid corpus for some of the UDHR
languages is dominated by religious texts.

\section{\genericname v1.0}\seclabel{glotcc}
The process of \genericname creation is similar to other
pipelines for large-scale web corpus
creation~\citep{kudugunta2024madlad}. More specifically, we
use Ungoliant  \cite{abadji2021ungoliant}, 
the OSCAR pipeline
\cite{OrtizSuarezSagotRomary2019}.
Instead of 
OSCAR's \fasttext LID (which 
detects 176~languages), we use \glotlid v3.0.
Recall that this change also includes the zxx and und labels that are part of \glotlid's label set and
support robust noise removal
(\secref{glotlid}). We further provide extensions for
content classes (\secref{categories}) and quality warnings
and filtering (\secref{quality}, \secref{filters}).
We also perform replacement of personally identifiable information  (\secref{pii})
for
\genericname. We distribute our forked pipeline
under the Apache 2.0 license, the same as the Ungoliant
license. \genericname is licensed under \cc terms: 
\href{https://commoncrawl.org/terms-of-use}{\path{commoncrawl.org/terms-of-use}}.

For this
paper,
we ran the pipeline on the \cc CC-MAIN-2024-10 snapshot
in its entirety and on parts of
CC-MAIN-2023-40 and CC-MAIN-2023-50. We refer to the corpus produced by this process as
\genericname\ v1.0 (or \genericname for short).

\input{tables/glotcc_geo_dist}

\genericname v1.0 contains data (subcorpora) for
\numlanguagesexact LID labels (i.e., language-script pairs
such as ``rus-Cyrl''). 
\tabref{glotcc_geo_dist} shows their
geographic distribution for
Glottolog~\citep{hammarstrom2024glottolog}
macroareas.\footnote{ ``Papunesia'' refers to ``Insular''
Southeast Asia and Oceania, excluding Australia.}  Based on
the Wikipedia list of ISO 639-3
codes,\footnote{\href{https://en.wikipedia.org/wiki/Codes_for_constructed_languages}{\path{wikipedia.org/wiki/Codes_for_constructed_languages}}}
we also add 12 constructed languages (``Constructed'').
As
reflected in Table~\ref{tab:glotcc-dist}, \genericname{}
v1.0 considerably increases language coverage compared to
OSCAR 23.01, especially for minority languages. The number of
languages with more than $10^{2}$ documents in
\genericname{} is 145+89+52+29+22+12=349
(vs 132 for OSCAR), and with more
than $10$ documents is 349+360=709 (vs 142 for OSCAR). This
language coverage
can be easily increased by applying \glotlid{} on
more \cc{} snapshots.
One reason for \genericname's better coverage could be that
we lose less minority language content via contamination
\citep{blevins-zettlemoyer-2022-language} to 
``large''
languages.
\genericname's Wikipedia percentage
is highest (.2658, .2940)
for languages with a document count between
$10^{2}$ and $10^{4}$. Many \genericname languages
with $\leq 10^{2}$ documents  come from
religious websites (.4441, .4285).
\genericname's coverage 
of languages
with more than $10^7$ documents
is lower than OSCAR's
because
we apply more
cleaning filters.

\input{tables/glotcc_dist}

\subsection{Annotating documents with content classes\seclabel{categories}}
Similar to OSCAR 23.01, we use the UT1 blocklist~\citep{ut1} to
classify websites into different content classes such as ``adult'' and ``blog''.
The main utility of these UT1-based filters is to warn about
potential adult content websites, with 3.5 million domains
labeled as adult in UT1. 

We add to UT1 two additional content classes: ``wikipedia'' and ``religious''.\footnote{We compile the religious
category based on popular websites like \url{jw.org} and \url{ebible.com}.
While we believe it to be a helpful indicator for
the user, it is not perfect, neither in terms of recall
(especially for non-Christian content) nor precision (many
religious websites also contain non-religious content).}
The religious content class is important to assess how
domain-specific a particular language's corpus is --
for some minority languages almost all web content
is religious~\citep{kudugunta2024madlad}. Following
OSCAR 23.01, we do not remove any content classes from
\genericname and instead leave that decision
(e.g., removal of adult content) to the user, mainly because of UT1 false positives~\citep{abadji2022towards}.

\subsection{Quality warnings}
\seclabel{quality}
We add new
quality warnings adopted from prior work on data
cleaning and web crawling~\citep{raffel2020exploring,
  rae2021scaling, abadji2022towards,
  imanigooghari-etal-2023-glot500,
  kargaran-etal-2024-glotscript-resource,
  kudugunta2024madlad}.
Specifically, we provide the following quality warnings.\footnote{In the following discussion, we
use the terms ``line'' and ``sentence'' interchangeably.}

\textbf{Tiny}: The document has a small number of
lines.
Following \citep{raffel2020exploring}, we use a threshold of three lines.

\textbf{Short sentences}: The document has a high number ($\geq$ 50\%) of short lines~\citep{abadji2022towards}.

\textbf{Header and footer}: \cc
contains boilerplate extracted from headers and footers.
We give a \textit{header}
(resp.\ \textit{footer})
warning if
short lines occur at the start (resp.\ end) of the
document
\citep{abadji2022towards}.

\textbf{Inconsistent}: Ungoliant applies LID at both 
document  and line levels. If $\geq$60\% of 
lines do not match the document-level label, we mark the
document as \textit{LID-inconsistent}. If
$\geq$10\% of the script content is incompatible
with the label
predicted by
LID, we mark the document as
\textit{script-inconsistent}~\citep{kargaran-etal-2024-glotscript-resource}. 

\textbf{List case}: Flag sentences with $\geq$50\% of  tokens beginning
with a capital letter~\citep{kudugunta2024madlad}. Some
scripts like Hani lack capital letters. However, since Chinese is written without spaces, we can still find lists  by identifying portions of the text that are shorter than five characters and are surrounded by spaces.

\textbf{Technical characters}: Fires when $\geq$20\% of  characters are
numbers/punctuation \citep{kudugunta2024madlad}. The warning
\textit{script-inconsistent} is also raised here since
$\geq$10\% is not written in the main script.

\textbf{Cursed regex}: These are substrings and regexes from~\citep{kudugunta2024madlad} used for identifying noisy and questionable content.

\textbf{Repetition}: Repetition of words and bigrams
indicates poor quality \citep{rae2021scaling}.
We give the warning \textit{repetition}
if a sentence has
$>$20 words and either  $>$50\% of the
words or $>$20\% of the bigrams are repetitive.
This  heuristic
cannot be applied to languages without word boundaries.

\textbf{Long word}:
This warning applies if there is a word with more than 100
characters;
see \cite{kudugunta2024madlad}.

\textbf{Lorem ipsum}: If the text contains the placeholder ``lorem ipsum''~\citep{raffel2020exploring}, we flag it as \textit{lorem ipsum}.

\textbf{Policy}: Many texts have boilerplate policy
notices~\citep{raffel2020exploring}.
We flag the text as \textit{policy}
if it contains any of the following: ``terms of use'', ``privacy policy'', ``cookie policy'', ``uses cookies'', ``use of cookies'', ``use cookies''.

\textbf{JS warning}: Many texts contain warnings that JavaScript should be enabled~\citep{raffel2020exploring}. If the text contains ``JavaScript'' or ``Javascript'', we flag it as \textit{js warning}.

\textbf{Curly bracket}: Curly braces are mostly used in programming languages, not natural language~\citep{raffel2020exploring}. If the text contains curly braces, we flag it as \textit{curly bracket}.

\textbf{Adult words}: This is a collection of pornographic keywords
from \citep{kudugunta2024madlad}, mostly for Chinese tokens.
We also found out that the GPT-4o
tokenizer~\citep{OpenAI2024} has many adult tokens and gambling terms, so we conducted a manual audit of Chinese tokens longer than 3
characters in this tokenizer's vocabulary 
and added them as additional keywords.

\subsection{Quality warning filters}\seclabel{filters}
We sample 20
sentences from three languages for each script: one
from the top 10\% of the language distribution (i.e., with
most data in \genericname),
one from the bottom 75\% and one from the remaining 15\%.
For each sentence, we determine whether it is
high-quality content in the target language.
For unknown languages, we check the website URL and
search for information about the language.
We find that the quality warnings generally indicate bad
content or erroneously assigned LID labels and therefore remove
sentences with quality warnings in \genericname. There are
two exceptions.

First, we ignore three warnings: \textit{short sentences}, \textit{header}, and \textit{footer}, because the LID label is correct for most of these documents, or in other cases, the warnings are false positives.
Second, 
for languages without
overt word
boundaries,\footnote{\href{https://en.wikipedia.org/wiki/Category:Writing_systems_without_word_boundaries}{\path{wikipedia.org/wiki/Category:Writing_systems_without_word_boundaries}}}
we keep documents with warnings
\textit{long word} and \textit{repetition}
as computing these warnings is nonsensical if the text is not
separated into words.
The warning metadata for these five warnings is kept in
\genericname in case users want to use it.

\subsection{Deduplication}\seclabel{dedup}

Following OSCAR 23.01, 
a hash is provided for
each document in \genericname. This hash is
computed by \texttt{py-tlsh}\footnote{\href{https://pypi.org/project/py-tlsh/}{\path{pypi.org/project/py-tlsh}}}
with hyperparameters  256 buckets and
3-byte checksums~\citep{abadji2022towards}. However, we do
not distribute deduplication as part of the pipeline, because 
deduplication is costly~\citep{gao2020pile,
  abadji2022towards} and
hashing algorithm and
hyperparameters are application-dependent.

\subsection{Personally identifiable information replacement}\seclabel{pii}
We replace two types of personally identifiable information (PII):
email addresses and public network IP addresses. We do
not replace phone numbers due to the high false
positive rate of regex patterns. We use the PII process implemented by DataTrove~\citep{penedo2024datatrove} (see also FineWeb~\citep{penedo2024fineweb}). Email addresses are replaced with ``email@example.com'' or ``firstname.lastname@example.com,'' and public network IP addresses are replaced with one of six IP addresses: ``22.214.171.124,'' ``126.96.36.199,'' ``188.8.131.52,'' ``184.108.40.206,'' ``220.127.116.11,'' or ``18.104.22.168,'' which, at the time of corpus creation, were unresponsive to pings.

\subsection{Wall time}\seclabel{walltime}

We calculate the pipeline's wall time, considering only the LID, without accounting for quality warning filters, the PII process, or any infrastructure-related bottlenecks. Suppose the LID throughput is, on average, \( T_S \) sentences per second and \( T_D \) documents per second, and we can run \( P \) parallel jobs. We have \( D \) documents to annotate, each containing an average of \( S \) sentences. For each document, in addition to running the LID on the entire document, we also run it on each individual sentence. The estimated processing time in hours is given by:

\[
\text{Wall time (hours)} = \frac{D}{3600 \times P} \times \left(\frac{S}{T_S} + \frac{1}{T_D}\right)
\]

The processing time on Intel Xeon E7-8857 3GHz CPUs with \( P = 48 \) for one Common Crawl snapshot (\( D = 3.16 \times 10^9 \) for CC-MAIN-2024-10) is estimated, using the given values of \( T_S = 1379 \), \( T_D = 245 \), and \( S = 20 \), to be approximately \( 340 \) hours.

\subsection{Self-audit quality review}\seclabel{self-audit}
We perform a
self-audit
of \genericname-V1.0. We have two motivations. First,
following \citep{kreutzer2022quality}, we want to ensure
that the target language metadata are correct and
that there are no systematic issues. Second, we intend to
develop additional filters to clean the corpus for future
releases. The bottleneck in such an audit is the difficulty
of finding native speakers for each language. Therefore,
following \citep{kudugunta2024madlad}, we conduct the audit
by providing high-level comments on the data quality and
identifying the language of the data by looking for language
clues.

We randomly select 653 languages. We sample 20
pages from each (or  all if there are fewer than 20)
and check the validitiy of \genericname's LID label.
Additionally, we 
analyze common errors
and provide high-level comments.

We follow these guidelines in conducting the audit:

\begin{itemize}[nolistsep]
\item For unknown languages, we inspect the URL
  and visit the webpage to find language clues such
  as language codes (especially in the URL), the \texttt{lang} attribute inside the html tag, country flag, contact address and the name of the language in the text. Otherwise, we search for sentences on the web to consult similar webpages related to that sentence.
\item If the corpus contains noise but the noise appears filterable, we leave a high-level note detailing the noise and how it can be filtered.
\item We report the percentage of in-language sentences for each audited language.

\end{itemize}

\textbf{Overall results.} Out of 653 audited languages, we find that, with a macro-average score of 0.93 and a median score of 1.0, the data is in-language. 
During the audit, for some languages, we couldn't determine the correct language; therefore, we do not consider those languages in the audited set.
We find 10 out-of-model
languages:
aon (Bumbita Arapesh, Torricelli),
bkx (Baikeno, Malayo-Polynesian),
gup (Gunwinggu, Arnhem, Northern Australia),
ibl (Ibaloi, Philippine), 
kpo (Ikposo, Atlantic-Congo),
mcr (Menya, Trans-New Guinea),
mge (Mango, Nilo-Saharan),
mrh (Mara Chin, Sino-Tibetan),
sxw (Saxwe Gbe, Atlantic-Congo) and
tao (Yami, Malayo Polynesian).
We plan to include these
languages in \glotlid v4.0.
There are also errors that neither the
LID nor the filters captures. For example, repetitive
n-grams in list-like content, such as those at the start or
end of words from websites like
\href{https://anagrams.app}{\path{anagrams.app}}.
Based on these audits, we published a more clean version of \genericname-V1.0 to the community.

\subsection{Evaluation of LID within the pipeline}
We compare the NLLB LID~\citep{nllbteam2022language} and
\glotlid within the context of the Ungoliant pipeline. For this comparison, we run the 
pipeline
in the exact same configuration, except that we
use NLLB LID for one run and \glotlid for the other.
We look at a random sample of minority language pages
for which the two pipelines make different predictions.

In more detail,
we select a subset of size $\approx$1\% (a prefix)
of the latest \cc snapshot (CC-MAIN-2024-18) and 
run the \genericname pipeline on it, once using \glotlid as the LID
and once using the NLLB LID as the LID. This
produces a corpus that is analogous to
\genericname-v1.0, except it is based on (a
prefix of) a different \cc snapshot.
We count the number of times that an LID label (e.g., rus-Cyrl)
is assigned by either \glotlid or NLLB LID in this ``filter'' subset; we refer to this
number as $n_l$ for label $l$. We restrict the comparison to
those labels that occur $n_l \leq 10$ times because our
main focus is minority languages, not ``large'' languages
that are already well covered by existing resources.
We further only consider pages for which \glotlid and
NLLB LID
disagree. This gives us a final set of 260 pages.
Note that this set of 260 pages would correspond to
$100 \times 100 \times 260$ = 2,600,000 pages, had we run the pipeline on the
entire \cc. (We take a prefix of size 1\% of the latest
snapshot and there have been about 100 \cc snapshots so far.)

\input{tables/nllb}

\tabref{nllb} reports a comparison of \glotlid and NLLB LID for a
random sample of 20 of the 260 pages; see codebase for
a detailed report of the comparison.
``miss'' refers to cases where the LID did not make a
call.
We see that \glotlid is correct
13 times for pages that NLLB LID treated incorrectly (3
misclassified pages, 10 ``miss'' pages) and incorrect
5 times for correct decisions by NLLB LID (3 incorrect calls
and 2 incorrect ``miss'' decisions).
On 2 pages,
both \glotlid and NLLB LID
make incorrect decisions.

The main reason for \glotlid's
clearly better performance is 
that it makes many
more calls than the prior state of the art, without losing
overall accuracy. This is in keeping with the substantial
expansion of GlotLID's label set (more than 2,000) compared
to prior work.

\section{Related work}

\subsection{LID}

There is a wealth of resources to perform LID, but not
all of them meet the requirements of a good LID for minority
corpus creation~\citep{kargaran-etal-2023-glotlid}.
All of the following cover at most 218 languages:
CLD2~\citep{mccandless2010accuracy},
Equilid~\citep{jurgens-etal-2017-incorporating},
Langdetect~\citep{shuyo2010language},
langid.py~\citep{lui-baldwin-2012-langid},
OpenLID~\citep{burchell-etal-2023-open}, NLLB
LID~\citep{nllbteam2022language}, FastText
LID~\citep{bojanowski-etal-2017-enriching},
CLD3~\citep{botha-etal-2017-natural,salcianu2018compact},
and HeLI-OTS~\citep{jauhiainen-etal-2022-heli}. Some LIDs are not open-source, e.g., those
published by \citet{caswell-etal-2020-language,
  bapna2022building,
  kudugunta2024madlad}. whatlang~\citep{la-strings,
  brown2012finding} and idNet~\citep{dunn2020mapping} are
two broad-coverage LIDs that meet many other requirements
but are hard to use in many practical scenarios due to
software issues and lack of maintenance. Franc~\citep{franc}
is a character 3-gram LID with support for 400+ languages;
however, it does not provide well-calibrated
probabilities. Another LID with similar properties and
support for 1600+ languages is
FUN-LangID~\citep{Fun_LangID};
however, according to the developers, this model is not the
best for high performance on F1/FPR.
AfroLID~\citep{adebara-etal-2022-afrolid}, which 
covers African languages, is a Transformer
architecture and less efficient than its competitors.
A strong requirement for LIDs is
their effective applicability for very large corpora,
not least
for ecological
reasons. GeoLID~\citep{dunn-edwards-brown-2024-geographically-informed}
meets most of the requirements; it supports 900+ languages
and the architecture is based on FastText. However, it needs
geographic prior information, which makes it more suitable
for social media such as X (Twitter) that provide such
geographic priors.

\subsection{Multilingual corpora}
Much work has been done on mining multilingual corpora from
the web.
\citet{xue-etal-2021-mt5} introduce mC4, a general
101-language web domain corpus,  to train the mT5
model. Similarly, \citet{conneau-etal-2020-unsupervised}
introduce CC-100 based on the CC-Net
repository~\citep{wenzek2020ccnet} to train the XLM-R
model. The OSCAR corpus~\citep{abadji2022towards} 
supports 150+ languages. The mC4 pipeline
uses CLD3, and CC-Net and OSCAR use FastText LID.
\citet{nllbteam2022language} mine
an internal corpus from CommonCrawl with 200+ languages using
NLLB LID. The MADLAD-400
corpus~\citep{kudugunta2024madlad} is another mined corpus
with 450+ languages using an internal 500-language coverage
LID and pipeline. The most closely related work to ours is
by~\citet{bapna2022building}, who create an internal
corpus of 1500+ languages using an internal 1600+ LID and
pipeline. There are also high-quality pipelines that focus
on creating corpora mostly for English,
including Dolma~\citep{soldaini2024dolma},
RedPajama-Data-v2~\citep{together2023redpajama}, DCLM~\citep{li2024datacomp} and
FineWeb~\citep{penedo2024fineweb}. There is also some work
that does not only mine the web, but builds mulitlingual data upon other
datasets by compiling multiple data sources, such as the
ROOTS corpus~\citep{laurencon2023bigscience}, a community-built
dataset that contains 46 languages.
CulturaX~\citep{nguyen2023culturax} combines mC4 3.1.0 and different OSCAR versions and applies additional filters, resulting in 160+ languages.
The Glot500
corpus~\citep{imanigooghari-etal-2023-glot500} covers 500+
languages (400+ open-access), mostly based on prior work in academia.
Serengeti~\citep{adebara-etal-2023-serengeti}
also introduces an internal dataset of 500+ African
languages derived from religious texts, news and academia.

To the best of our knowledge, \genericname is the first
open-access corpus that is based on an open pipeline (including open-access
LID) and  that passes the threshold of 160 languages (\eg CulturaX~\citep{nguyen2023culturax})  in
web-mined corpora; in fact, as we show we cover more
than a thousand languages. As our statistics show, we find many
languages on the web that have never before been part of a
web-mined corpus.

\section{Limitations}\seclabel{limitations}

\textbf{Use cases:} Due to certain filtering steps (\eg curly bracket filter), the \genericname
likely does not contain much math/code content. It is
advisable to supplement \genericname  with
math/code data if this is the intended use
case. Due to the use of LID and script inconsistency filters as indicators of noise, the documents that exist in the corpus are more monolingual than multilingual or code-switched~\citep{kargaran2024masklid} compared to when these filters are not used. Additionally, since we did not customize the processing
for each website, some sources, such as
Wikipedia, may have better formatting in the original than
in \genericname.

\textbf{Errors:} Although we have good in-language content, \genericname still exhibits many types of errors and noise,
including misclassification and out-of-model cousins.

\textbf{Model training:} We did not train any language model to better justify the significance of GlotCC, as this would be pointless unless we also evaluate them. Evaluating language models requires evaluation data that we mostly do not have for minority languages (see \citep{kargaran2024mexa, ahuja-etal-2024-megaverse, ahuja-etal-2023-mega, imanigooghari-etal-2023-glot500}). At this stage, this endeavor seems unrealistic. We believe that identifying relevant data and performing LID is a crucial first step in that direction.

\section{Conclusion}

We introduce a document-level, general domain web corpus covering more than \numlanguages languages. We open-source the entire pipeline, including a better language identification model, which is more robust in the corpus creation task in terms of noise handling, unseen writing systems, and a broad coverage of languages to reduce the chance of encountering unknown languages. We filter the created corpus and perform a self-audit to ensure the created corpus is clean.

We hope the creation of such a corpus and pipeline will benefit language technologies, enabling the inclusion of more minority languages. For future work, we plan to extend this corpus to additional \cc snapshots.

\section{Ethics statement}\seclabel{ethics}

The advancement of NLP technologies has primarily been
constrained to languages for which resources are available
in good quality and quantity. Many of the
world's minority languages face a significant barrier due to
the scarcity of high-quality general data sources, making it
difficult to develop NLP tools for these languages.
Despite concerns that an ``extractive'' approach to NLP
often does not benefit the affected communities
\citep{bird2020decolonising}, it is still an important goal
of both computational and theoretical linguistics to have as
good a representation of minority languages in available
web-based corpora  as is permitted by the licenses of
content available on the web.
By
providing a cleaned corpus like \genericname, we take
initial steps towards including a diverse range of languages
in  NLP. Despite our strong focus on   filtering,
the cleaning of \genericname is constrained by the
lack of tools for filtering out undesirable content and noise -- such as
adult content, personal information, lists --
and this is a considerable problem for a subset of
languages, including languages without explicit word boundaries.
Therefore, we
recommend users carefully evaluate their specific use case
before using \genericname.

\begin{ack}
We would like to thank anonymous reviewers.
This work was funded by Deutsche Forschungsgemeinschaft (project SCHU 2246/14-1).
\end{ack}

\bibliographystyle{plainnat}
\bibliography{main}

\newpage
\section*{Checklist}

\input{checklist}

\newpage
\appendix

\section{Appendix}

\subsection{\glotlid model card}\applabel{model_card}

We provide the \glotlid v3.0 model card based on the model card template introduced by \citet{mitchell2019model}.

\input{modelcard}

\newpage

\subsection{\genericname datasheet}\applabel{datasheet}

We provide the \genericname v1.0 datasheet based on the datasheet template introduced by \citet{gebru2021datasheets}.

\input{datasheet}

\end{document}

%% file: tables/glotlid_hyperparams.tex
\setlength{\textfloatsep}{5pt}
\begin{figure}[h]
\small
 \centering
 \captionof{table}{\small{\glotlid v3.0 training hyperparameters}}
\begin{tabular}{llc}
\toprule
{Argument} & {Description} & {Value} \\ \midrule
{-minCount} & {Minimal number of word occurrences} & {1000}  \\
{-minCountLabel} & {Minimal number of label occurrences} & {0} \\
{-wordNgrams} & {Max length of word ngram} & {1} \\
{-bucket} & {Number of buckets} & {10$^{6}$} \\
{-minn} & {Min length of char ngram} & {2} \\
{-maxn} & {Max length of char ngram} & {5} \\
{-loss} & {Loss function} & {softmax} \\
{-dim} & {Size of word vectors} & {256} \\
{-epoch} & {Number of epochs} & {1} \\
{-lr} & {Learning rate} & {.8} \\
\bottomrule
\end{tabular}
\tablabel{hyperparameters}
\end{figure}

%% file: tables/glotlid_abs.tex
\setlength{\textfloatsep}{5pt}
\begin{wrapfigure}{r}{0.49\columnwidth}\vspace{-5mm}
 \centering
 \captionof{table}{\small{Performance of GlotLID v3.0}}
\begin{tabular}{lrcc}
\toprule
{Benchmark} & {\# Labels} & {F1\,$\uparrow$} & {FPR\,$\downarrow$}\\ \midrule
{GlotTest} & {2102} & {0.991} & {0.000003}\\
{UDHR} & {371} & {0.882} & {0.000298} \\
{FLORES-200} & {199} & {0.967} & {0.000161} \\ \bottomrule
\end{tabular}%
\label{tab:lid_performance} 
\end{wrapfigure}

%% file: tables/glotcc_geo_dist.tex
\setlength{\textfloatsep}{5pt}
\begin{wrapfigure}{r}{0.4\columnwidth}\vspace{-4.5mm}
\small
 \centering
 \captionof{table}{\small{Geographic distribution of languages in \genericname.}}
\begin{tabular}{lr}
\toprule
{Macroarea} & {\# Labels} \\ \midrule
{Eurasia} & {395} \\
{Papunesia} & {380} \\
{Africa} & {252} \\
{North America} & {123} \\
{South America} & {97} \\
{Australia} & {16} \\
{Constructed} & {12} \\ \bottomrule
\end{tabular}%
\label{tab:glotcc_geo_dist} 
\end{wrapfigure}

%% file: tables/glotcc_dist.tex
\begin{table}[h!]
\centering
\caption{\label{tab:glotcc-dist} Partition statistics for
  OSCAR 23.01 and \genericname-v1.0.
Each partition is defined as:
$10^{J} > \text{\# documents per language} \geq 10^{I}$
where
$ 0 \leq I \leq 7 $,
$ 1 \leq J \leq 9 $.
}

\resizebox{0.99\columnwidth}{!}{
\begin{tabular}{@{}llccccccccc@{}}
\toprule
\multirow{2}{*}{\{$I$, $J$\}}  &
\multirow{2}{*}{Corpus Version} & 
\multirow{2}{*}{\# Languages} &
\multicolumn{2}{c}{\# Documents} & \multicolumn{2}{c}{\# Lines} & \multicolumn{2}{c}{\# Words} & \multicolumn{1}{c}{\# Religious} & \multicolumn{1}{c}{\# Wikipedia}   \\ \cmidrule(l){4-11} 
                                 & & & Total           & Median          & Total           & Median          & Total         & Median  & Total pct.
  & Total pct. \\ \midrule
\multirow{2}{*}{\{7, 9\}} & OSCAR 23.01 & 24 & 2.7B & 34.4M & - & - & 1.0T & 12.6B & - & - \\
& \genericname-v1.0 & 12 & 579.5M & 22.7M & 15.1B & 780.8M & 436.4B & 17.0B & 0.0001 & 0.0009 \\                      \midrule     
\multirow{2}{*}{\{6, 7\}} & OSCAR 23.01 & 23 & 80.0M & 2.4M & - & - & 27.6B & 738.8M & - & - \\
& \genericname-v1.0 & 22 & 92.2M & 3.8M & 3.0B & 122.1M & 67.8B & 2.4B & 0.0001 & 0.0044 \\
\midrule
\multirow{2}{*}{\{5, 6\}} & OSCAR 23.01 & 25 & 9.3M & 262.7K & - & - & 3.2B & 82.4M & - & - \\
& \genericname-v1.0 & 29 & 10.7M & 334.8K & 305.4M & 9.1M & 6.9B & 195.7M & 0.0001 & 0.0219 \\                      \midrule     
\multirow{2}{*}{\{4, 5\}} & OSCAR 23.01 & 26 & 919.7K & 25.2K & - & - & 212.0M & 5.4M & - & - \\
& \genericname-v1.0 & 52 & 1.9M & 29.6K & 55.1M & 714.4K & 1.3B & 17.9M & 0.0005 & 0.0922 \\
\midrule
\multirow{2}{*}{\{3, 4\}} & OSCAR 23.01 & 14 & 60.1K & 3.6K & - & - & 10.1M & 315.7K & - & - \\
& \genericname-v1.0 & 89 & 338.7K & 2.7K & 8.2M & 52.2K & 223.9M & 1.4M & 0.0029 & 0.2658 \\
\midrule
\multirow{2}{*}{\{2, 3\}} & OSCAR 23.01 & 20 & 8.6K & 400 & - & - & 772.3K & 13.4K & - & - \\
& \genericname-v1.0 & 145 & 53.9K & 326 & 1.4M & 6.5K & 39.3M & 192.6K & 0.0606 & 0.2940 \\
\midrule
\multirow{2}{*}{\{1, 2\}} & OSCAR 23.01 & 10 & 368 & 36 & - & - & 13.6K & 431 & - & - \\
& \genericname-v1.0 & 360 & 11.5K & 24 & 245.0K & 460 & 11.3M & 20.5K & 0.4441 & 0.1044 \\
\midrule
\multirow{2}{*}{\{0, 1\}} & OSCAR 23.01 & 10 & 44 & 4 & - & - & 21.5K & 67 & - & - \\
& \genericname-v1.0 & 566 & 1.7K & 2 & 41.5K & 26 & 1.7M & 1.2K & 0.4285 & 0.0285 \\
\midrule
\midrule
\multirow{2}{*}{\{0, 9\}} & OSCAR 23.01 & 152 & 2.8B & 69.7K & - & - & 1.1T & 14.5M & - & - \\
& \genericname-v1.0 & 1275 & 684.7M & 14 & 18.5B & 254 & 512.6B & 11.6K & - & - \\
\bottomrule
\end{tabular}
}
\end{table}

%% file: tables/nllb.tex
\def\mycolsep{0.08cm}
\setlength{\textfloatsep}{5pt}
\begin{wrapfigure}{r}{0.53\columnwidth}\vspace{-4.5mm}
 \centering
   \captionof{table}{\small{Comparison of \glotlid and NLLB on a random
  subset of 20 pages from minority languages\tablabel{nllb}}}
   \begin{tabular}{l||lc@{\hspace{\mycolsep}}c@{\hspace{\mycolsep}}c@{\hspace{\mycolsep}}c}
    & \multicolumn{1}{c|}{NLLB$\rightarrow$} & correct&error&miss\\\hline\hline
  \glotlid$\downarrow$ &&\\\cline{1-1}
  correct & & 0     &3        &10   &\\
  error&& 3     &2        &0   &\\
  miss     && 2     & 0       & -  &\\
\end{tabular}
\end{wrapfigure}

%% file: checklist.tex
\begin{enumerate}

\item For all authors...
\begin{enumerate}
  \item Do the main claims made in the abstract and introduction accurately reflect the paper's contributions and scope?
    \answerYes{}
  \item Did you describe the limitations of your work?
    \answerYes{See \Secref{limitations}.}
  \item Did you discuss any potential negative societal impacts of your work?
    \answerYes{See \Secref{ethics}.}
  \item Have you read the ethics review guidelines and ensured that your paper conforms to them?
    \answerYes{}
\end{enumerate}

\item If you are including theoretical results...
\begin{enumerate}
  \item Did you state the full set of assumptions of all theoretical results?
    \answerNA{}
	\item Did you include complete proofs of all theoretical results?
    \answerNA{}
\end{enumerate}

\item If you ran experiments (e.g. for benchmarks)...
\begin{enumerate}
  \item Did you include the code, data, and instructions needed to reproduce the main experimental results (either in the supplemental material or as a URL)?
    \answerYes{See the Abstract, Appendices \ref{sec:model_card} and \ref{sec:datasheet}.}
  \item Did you specify all the training details (e.g., data splits, hyperparameters, how they were chosen)?
    \answerYes{See \secref{evaluation}}
	\item Did you report error bars (e.g., with respect to the random seed after running experiments multiple times)?
    \answerYes{\genericname does not depend on any random seed. For \glotlid see \secref{evaluation}}
	\item Did you include the total amount of compute and the type of resources used (e.g., type of GPUs, internal cluster, or cloud provider)?
    \answerYes{See \secref{walltime}}
\end{enumerate}

\item If you are using existing assets (e.g., code, data, models) or curating/releasing new assets...
\begin{enumerate}
  \item If your work uses existing assets, did you cite the creators?
    \answerYes{}
  \item Did you mention the license of the assets?
    \answerYes{See \Appref{model_card} and \Appref{datasheet} (or \Secref{glotcc}).}
  \item Did you include any new assets either in the supplemental material or as a URL?
    \answerYes{}
  \item Did you discuss whether and how consent was obtained from people whose data you're using/curating?
    \answerYes{See \Appref{datasheet} (or \Secref{glotcc}).}
  \item Did you discuss whether the data you are using/curating contains personally identifiable information or offensive content?
    \answerYes{See \Appref{datasheet}, Sections \ref{sec:pii} and \ref{sec:ethics}.}
\end{enumerate}

\item If you used crowdsourcing or conducted research with human subjects...
\begin{enumerate}
  \item Did you include the full text of instructions given to participants and screenshots, if applicable?
    \answerYes{See \secref{self-audit}.}
  \item Did you describe any potential participant risks, with links to Institutional Review Board (IRB) approvals, if applicable?
    \answerNA{}
  \item Did you include the estimated hourly wage paid to participants and the total amount spent on participant compensation?
    \answerNo{The audits are performed by the authors.}
\end{enumerate}

\end{enumerate}

%% file: modelcard.tex
\fbox{\begin{minipage}{43em}

\small

\paragraph{Model details}

\begin{itemize}[leftmargin=*]
    \item Person or organization developing model: \textit{LMU Munich and Sorbonne Université}
    \item Model Date: \textit{April 18, 2024}
    \item Model Types: \textit{Language identification and language modeling.}
    \item Model Access: \github \href{https://github.com/cisnlp/GlotLID}{\path{github.com/cisnlp/GlotLID}} and \huggingface \href{https://huggingface.co/cis-lmu/glotlid}{\path{hf.co/cis-lmu/glotlid}}
    \item Information about training algorithms, parameters, fairness constraints or other applied approaches, and features: \textit{Provided in the \Secref{glotlid}}
    \item Paper: \begin{itemize}
        \item V3: \textit{Kargaran et al, \genericname: An Open Broad-Coverage \cc Corpus and Pipeline for Minority Languages, NeurIPS, 2024}
    \item V1: \textit{Kargaran et al, \glotlid: Language Identification for Low-Resource Languages, EMNLP, 2023}
    \end{itemize}
    \item License: \textit{Apache License 2.0}
    \item Contact: \textit{amir@cis.lmu.de}
\end{itemize}

\paragraph{Intended use} 

\begin{itemize}[leftmargin=*]
    \item Primary intended uses: \textit{Language identification on over 2000 linguistic labels.}
    \item Primary intended users: \textit{Research community.}
    \item Out-of-scope use cases: \textit{The model is trained on general domain data and might not perform well with short sentences or domain-specific data like the medical domain.}
\end{itemize}

\paragraph{Factors} 

\begin{itemize}[leftmargin=*]
    \item \textit{The classification quality of this model varies based on language, as some languages are easy to distinguish while others are challenging with n-gram models.}
\end{itemize}

\paragraph{Metrics}

\begin{itemize}[leftmargin=*]
    \item \textit{We use F1 score and FPR (false positive rate), two widely used metrics in the language identification literature, for our evaluations.}
\end{itemize}

\paragraph{Ethical considerations} 

\begin{itemize}[leftmargin=*]
    \item \textit{We acknowledge that this model has a high error rate for some of the languages. This means that there is a potential risk of excluding minority languages during the collection and processing of NLP corpora.}
\end{itemize}

\paragraph{Training data} 

\begin{itemize}[leftmargin=*]
    \item \textit{We train this model using openly available (but not necessarily freely redistributable) datasets, including resources previously published by researchers, publishers, and translators. The complete list of data sources is available on \\ \github \href{https://github.com/cisnlp/GlotLID/blob/main/sources.md}{\path{github.com/cisnlp/GlotLID/blob/main/sources.md}}.}
\end{itemize}

\paragraph{Evaluation data} 

\begin{itemize}[leftmargin=*]
    \item \textit{For evaluation, we used GlotTest, UDHR, and Flores-200 as described in \secref{lid_evaluation_data}.}
\end{itemize}

\paragraph{Caveats and recommendations}

\begin{itemize}[leftmargin=*]
    \item \textit{We note that if there is a setup where the list of languages is known, then the model can limit its predictions to that set of languages. This implies that languages not included in the set will be excluded from the softmax computation. We have provided the code for limiting this on \github \href{https://github.com/cisnlp/GlotLID}{\path{github.com/cisnlp/GlotLID}}.}
\end{itemize}

\end{minipage}}

%% file: datasheet.tex

\small

\paragraph{Motivation}

\begin{enumerate}[leftmargin=*]
    \item For what purpose was the dataset created? (Was there a specific task in mind? Was there a specific gap that needed to be filled? Please provide a description.)
    \textit{We created \genericname as a general web-crawled document-level dataset covering \numlanguagesexact languages, with the purpose of breaking barriers by providing training data for language technologies to include a broader range of languages.}
    \item Who created this dataset and on behalf of which entity? \textit{Amir Hossein Kargaran$^{\clubsuit}$,
    François Yvon$^{\spadesuit}$,
    Hinrich Schütze$^{\clubsuit}$
    ($^\clubsuit$LMU Munich,   
    $^\spadesuit$Sorbonne Université})
    \item Who funded the creation of the dataset? \textit{DFG (grant SCHU 2246/14-1).}
    \item Any other comments? \textit{None.}
\end{enumerate}

\paragraph{Composition}

\begin{enumerate}[leftmargin=*]
    \item What do the instances that comprise the dataset represent? \textit{Each instance is a filtered instance from \cc, with its language annotated by \glotlid.}
    \item How many instances are there in total? \textit{\genericname has 684.7M documents (18.5B lines, or 512.6B words) total across \numlanguagesexact languages.}
    \item Does the dataset contain all possible instances or is it a sample (not necessarily random) of instances from a larger set? (If the dataset is a sample, then what is the larger set? Is the sample representative of the larger set (e.g., geographic coverage)? If so, please describe how this representativeness was validated/verified. If it is not representative of the larger set, please describe why not (e.g., to cover a more diverse range of instances, because instances were withheld or unavailable).) \textit{\genericname is created from \cc and has been annotated by \glotlid, then filtered. To maintain a high level of in-language content, we have employed aggressive filtering, which may result in the exclusion of certain documents in a specific language within \cc.}
    \item What data does each instance consist of? \textit{Each instance is a raw text with accompanying metadata, including the timestamp, URL, quality warnings, category, tlsh hash, language identification probability, language identification consistency score, script consistency score, number of sentences, and content length.}
    \item Is there a label or target associated with each instance? If so, please provide a description. \textit{Yes, \genericname has a language label for each instance.}
    \item Is any information missing from individual instances? \textit{No.}
    \item Are relationships between individual instances made explicit? \textit{No.}
    \item Are there recommended data splits (e.g., training, development/validation, testing)? \textit{No.}
    \item Are there any errors, sources of noise, or redundancies in the dataset? \textit{
    Although \genericname have good in-language content, \genericname still exhibits many types of errors and noise, including misclassification and out-of-model cousins.}
    \item Is the dataset self-contained, or does it link to or otherwise rely on external resources (e.g., websites, tweets, other datasets)? (If it links to or relies on external resources, a) are there guarantees that they will exist, and remain constant, over time; b) are there official archival versions of the complete dataset (i.e., including the external resources as they existed at the time the dataset was created); c) are there any restrictions (e.g., licenses, fees) associated with any of the external resources that might apply to a future user? Please provide descriptions of all external resources and any restrictions associated with them, as well as links or other access points, as appropriate.) \textit{Yes.}
    \item Does the dataset contain data that might be considered confidential (e.g., data that is protected by legal privilege or by doctor-patient confidentiality, data that includes the content of individuals' non-public communications)? (If so, please provide a description.) \textit{It is possible as \genericname is a general web-crawled dataset.}
    \item Does the dataset contain data that, if viewed directly, might be offensive, insulting, threatening, or might otherwise cause anxiety? (If so, please describe why.) \textit{It is possible as \genericname is a general web-crawled dataset.}
    \item Does the dataset relate to people? \textit{It’s possible that some instances of \genericname mention and describe individuals.}
    \item Does the dataset identify any subpopulations (e.g., by age, gender)?  \textit{
    It’s possible that some instances of \genericname mention and describe people of certain subpopulations.}
    \item Is it possible to identify individuals (i.e., one or more natural persons), either directly or indirectly (i.e., in combination with other data) from the dataset? \textit{It’s possible that some instances of \genericname mention and describe individuals.}
    \item Does the dataset contain data that might be considered sensitive in any way (e.g., data that reveals racial or ethnic origins, sexual orientations, religious beliefs, political opinions or union memberships, or locations; financial or health data; biometric or genetic data; forms of government identification, such as social security numbers; criminal history)? \textit{It is possible as \genericname is a general web-crawled dataset.}
    \item Any other comments? \textit{None.}
\end{enumerate}

\paragraph{Collection}

\begin{enumerate}[leftmargin=*]
    \item How was the data associated with each instance acquired? \textit{    
    From \cc CC-MAIN-2024-10 snapshot eniterly and on parts from \cc CC-MAIN-2023-40 and CC-MAIN-2023-50.}
    \item What mechanisms or procedures were used to collect the data (e.g., hardware apparatus or sensor, manual human curation, software program, software API)? \textit{We annotated the \cc data using \glotlid and then filtered the documents to create \genericname.}
    \item If the dataset is a sample from a larger set, what was the sampling strategy? \textit{\genericname is a subset of CommonCrawl documents based on the \glotlid annotations and filtering steps.}
    \item Who was involved in the data collection process (e.g., students, crowdworkers, contractors) and how were they compensated (e.g., how much were crowdworkers paid)? \textit{For the audit, the authors inspected the dataset.}
    \item Over what timeframe was the data collected? (Does this timeframe match the creation timeframe of the data associated with the instances (e.g., recent crawl of old news articles)? If not, please describe the timeframe in which the data associated with the instances was created.) \textit{Each instance is followed by a UTC timestamp that shows the time of the crawl provided by \cc.}
    \item Were any ethical review processes conducted (e.g., by an institutional review board)? \textit{No.}
    \item Does the dataset relate to people? \textit{It's possible that some instances of \genericname mention and describe individuals.}
    \item Did you collect the data from the individuals in question directly, or obtain it via third parties or other sources (e.g., websites)? \textit{We collect the data from \cc.}
    \item Were the individuals in question notified about the data collection? \textit{No.}
    \item Did the individuals in question consent to the collection and use of their data? \textit{No.}
    \item If consent was obtained, were the consenting individuals provided with a mechanism to revoke their consent in the future or for certain uses? \textit{N/A.}
    \item Has an analysis of the potential impact of the dataset and its use on data subjects (e.g., a data protection impact analysis) been conducted? \textit{No.}
    \item Any other comments? \textit{None.}
\end{enumerate}

\paragraph{Preprocessing/cleaning/labeling}

\begin{enumerate}[leftmargin=*]
    \item Was any preprocessing/cleaning/labeling of the data done (e.g., discretization or bucketing, tokenization, part-of-speech tagging, SIFT feature extraction, removal of instances, processing of missing values)? (If so, please provide a description. If not, you may skip the remainder of the questions in this section.) \textit{We filter the data based on various quality warning filters.
}
    \item Was the "raw" data saved in addition to the preprocessed/cleaned/labeled data (e.g., to support unanticipated future uses)? \textit{We do not change the raw data; we only provide the metadata and determine which \cc instances to keep.}
    \item Is the software used to preprocess/clean/label the instances available? \textit{Yes, \github \href{https://github.com/cisnlp/GlotCC}{\path{github.com/cisnlp/GlotCC}}}
    \item Any other comments? \textit{None.}
\end{enumerate}

\paragraph{Uses}

\begin{enumerate}[leftmargin=*]
    \item Has the dataset been used for any tasks already? (If so, please provide a description.) \textit{No.}
    \item Is there a repository that links to any or all papers or systems that use the dataset? \textit{No.}
    \item What (other) tasks could the dataset be used for? \textit{\genericname can serve as a general training dataset for any language included in \genericname.}
    \item Is there anything about the composition of the dataset or the way it was collected and preprocessed/cleaned/labeled that might impact future uses? (For example, is there anything that a future user might need to know to avoid uses that could result in unfair treatment of individuals or groups (e.g., stereotyping, quality of service issues) or other undesirable harms (e.g., financial harms, legal risks) If so, please provide a description. Is there anything a future user could do to mitigate these undesirable harms?) \textit{
    Despite our strong focus on filtering, the cleaning of \genericname is constrained by the lack of tools for filtering out undesirable content and noise – such as adult content, personal information, lists – and this is a considerable problem for a subset of languages, including languages without explicit word boundaries. Therefore, we recommend users carefully evaluate their specific use case before using \genericname.}
    \item Are there tasks for which the dataset should not be used? (If so, please provide a description.) \textit{N/A.}
    \item Any other comments? \textit{None.}
\end{enumerate}

\paragraph{Distribution}

\begin{enumerate}[leftmargin=*]
    \item How will the dataset will be distributed (e.g., tarball on website, API, GitHub)? \textit{\genericname is made available through a Huggingface datasets \huggingface \href{https://huggingface.co/datasets/cis-lmu/GlotCC-V1}{\path{hf.co/datasets/cis-lmu/GlotCC-V1}}.}
    \item When will the dataset be distributed? \textit{June 2024.}
    \item Will the dataset be distributed under a copyright or other intellectual property (IP) license, and/or under applicable terms of use (ToU)? (If so, please describe this license and/or ToU, and provide a link or other access point to, or otherwise reproduce, any relevant licensing terms or ToU, as well as any fees associated with these restrictions.) \textit{
    LMU Munich hosts \genericname on Huggingface. We license the \genericname metadata under CC0 1.0 (public domain). However, the \genericname  is licensed under the terms of the \cc Terms of Use.
    }
    \item Have any third parties imposed IP-based or other restrictions on the data associated with the instances? \textit{\genericname is licensed under the terms of the \cc Terms of Use.}
    \item Any other comments? \textit{None.}
\end{enumerate}

\paragraph{Maintenance}

\begin{enumerate}[leftmargin=*]
    \item Who is supporting/hosting/maintaining the dataset? \textit{LMU Munich on Huggingface.}
    \item How can the owner/curator/manager of the dataset be contacted (e.g., email address)? \textit{Amir Hossein Kargaran (\texttt{amir@cis.lmu.de}).}
    \item Is there an erratum? (If so, please provide a link or other access point.) \github \href{https://github.com/cisnlp/GlotCC}{\path{github.com/cisnlp/GlotCC}}
    \item Will the dataset be updated (e.g., to correct labeling errors, add new instances, delete instances')? (If so, please describe how often, by whom, and how updates will be communicated to users (e.g., mailing list, GitHub)?) \textit{Yes, we maintain the data based on discussions on Huggingface, GitHub and those reported via email.}
    \item If the dataset relates to people, are there applicable limits on the retention of the data associated with the instances (e.g., were individuals in question told that their data would be retained for a fixed period of time and then deleted)? (If so, please describe these limits and explain how they will be enforced.) \textit{N/A.}
    \item Will older versions of the dataset continue to be supported/hosted/maintained? (If so, please describe how. If not, please describe how its obsolescence will be communicated to users.) \textit{Yes, any version of the data will be hosted for the sake of reproducibility of results for others using the dataset.}
    \item If others want to extend/augment/build on/contribute to the dataset, is there a mechanism for them to do so? (If so, please provide a description. Will these contributions be validated/verified? If so, please describe how. If not, why not? Is there a process for communicating/distributing these contributions to other users? If so, please provide a description.) \textit{We made the system to generate \genericname, including the pipeline, language identification model, and filters, available to the research community. This allows them to build upon it or customize it for their use case.}
    \item Any other comments? \textit{None.}
\end{enumerate}
